\documentclass{article}
\usepackage{spconf,amsmath,graphicx,booktabs}
\usepackage[utf8x]{inputenc}
\usepackage{hyperref}
\usepackage{multirow}

\title{Detecting Deception in Political Debates\\ Using Acoustic and Textual Features}
\name{Daniel Kopev$^1$, Ahmed Ali$^2$, Ivan Koychev$^1$, Preslav Nakov$^2$}
\address{
  $^1$Faculty of Mathematics and Informatics, Sofia University ``St Kliment Ohridski''\\
  $^2$Qatar Computing Research Institute, HBKU}

\begin{document}
%
\maketitle
\begin{abstract}
We present work on deception detection, where, given a spoken claim, we aim to predict its factuality. While previous work in the speech community has relied on recordings from staged setups where people were asked to tell the truth or to lie and their statements were recorded, here we use real-world political debates. Thanks to the efforts of fact-checking organizations, it is possible to obtain annotations for statements in the context of a political discourse as true, half-true, or false. Starting with such data from the CLEF-2018 CheckThat! Lab, which was limited to text, we performed alignment to the corresponding videos, thus producing a multimodal dataset. We further developed a multimodal deep-learning architecture for the task of deception detection, which yielded sizable improvements over the state of the art for the CLEF-2018 Lab task 2. Our experiments show that the use of the acoustic signal consistently helped to improve the performance compared to using textual and metadata features only, based on several different evaluation measures.
We release the new dataset to the research community, hoping to help advance the overall field of multimodal deception detection.
\end{abstract}
\begin{keywords}
deception detection, fact-checking, fake news, disinformation, computational paralinguistics, multimodality, political debates.
\end{keywords}

\section{Introduction}
\label{sec:intro}

Traditionally, news media have been the gate keepers of information, as they carefully selected what was appropriate to present to the public after double-checking it. Then, along came the Internet and social media, and the situation changed. Suddenly, anybody could publish and could share news online, which effectively stripped traditional media of their gate-keeping role, thus leaving the public largely unprotected against possible disinformation. Moreover, as mainstream media had to compete for user attention in social media against the new players, they had to play by the new rules, where speed was of utmost importance. This need for speed left very little time to double-check the information, and the overall quality of journalism started to degrade.

\noindent Furthermore, some politicians soon noticed that when it came to shaping public opinion, facts were secondary, and appealing to emotions worked better. 
Indeed, there are strong indications that false information was weaponized during the 2016 U.S. presidential campaign, thus marking the dawn of a Post-Truth Age. ``Fake news'', which can be defined as ``false, often sensational, information disseminated under the guise of news reporting'', became the word of the year in 2017, according to Collins Dictionary.
``Fake news'' thrive on social media thanks to the mechanism of sharing, which amplifies their effect.
Moreover, it has been shown that they spreads faster than real news~\cite{Vosoughi1146}. As they reach the same user multiple times, they could even be perceived as more credible.

As the problem became evident, a number of fact-checking initiatives have started, led by organizations such as FactCheck\footnote{\url{http://www.factcheck.org}}, Politifact\footnote{\url{http://www.politifact.com/}}, Snopes,\footnote{\url{http://www.snopes.com}}, Full Fact\footnote{\url{http://fullfact.org/}}, among others. Yet, manual fact-checking has proved to be a tedious task, and thus only a relatively small number of claims could be fact-checked. It is also slow: by the time the false information is debunked, it would have already traveled around the world and back and the harm could hardly be undone.

As a result, performing automatic fact-checking was proposed as a possible alternative. Ideally, journalists would have a hypothetical tool that would be able to detect and fact-check interesting claims in real time, i.e.,~as soon as the claims were made.
Thanks to the manual efforts by fact-checking organizations, it became possible to obtain annotations for the factuality of statements made in the context of a political discourse, which could be used to train automatic systems. 
Many fact-checking shared tasks, such as CLEF-2018 CheckThat Lab task 2~\cite{clef2018checkthat:task2}, used such real-world debates, but were limited to text. This is despite evidence from previous work that acoustics could help detect deception. Here we bridge this gap by aligning the debates to the corresponding videos, thus producing a multimodal dataset, the first of its kind: unlike previous work in the speech community, which has used recordings from staged setups where people were asked to tell the truth or to lie, here we use real-world political debates.

Our contributions can be summarized as follows:
\begin{itemize}
    \item We create the first multi-modal dataset for fact-checking the claims made in a political debate, which we release to the research community,\footnote{\small \url{http://github.com/fire0/detecting-deception-in-political-debates}} hoping to help advance research in multimodal deception detection.
    \item We further develop a multimodal deep-learning architecture for the task, which combines textual and acoustic information. Notably, we use no external information, which would not be available when a brand new claim is made for the first time.
    \item We show that the use of the acoustic signal helps to improve the factuality prediction, and we demonstrate sizable improvement over the state-of-the-art for task 2 of CLEF-2018 CheckThat! Lab.
\end{itemize}

The remainder of this paper is organized as follows: Section~\ref{sec:related} offers an overview of relevant related work. Section~\ref{sec:data} introduces our dataset.
Section~\ref{sec:features} presents our textual and acoustic.
Section~\ref{sec:experiments} describes our experiments and discusses the evaluation results.
Finally, Section~\ref{sec:conclusion} concludes and points to possible directions for future work.

\section{Related Work}
\label{sec:related}

Journalists, online users, and researchers are well-aware of the proliferation of false information, and thus topics such as credibility and fact-checking are becoming increasingly important.
At the claim-level, fact-checking and rumor detection have been primarily addressed using information extracted from social media, i.e.,~based on how users comment on the target claim \cite{Canini:2011,Castillo:2011:ICT:1963405.1963500,Ma:2015:DRU,ma2016detecting,PlosONE:2016,dungs-EtAl:2018:C18-1,kochkina-liakata-zubiaga:2018:C18-1}.
The Web has also been used as an information source \cite{mukherjee2015leveraging,Popat:2017:TLE:3041021.3055133,RANLP2017:factchecking:external,AAAI2018:factchecking,baly-EtAl:2018:N18-2,EMNLP2019:fauxtography}.

Previous work on deception detection \cite{Almela:2012:STD:2388616.2388619,Mihalcea:2009:LDE:1667583.1667679,NOdoi:10.1177/0146167203029005010} has relied on the use of the Linguistic Inquiry and Word Count (LIWC)\footnote{\url{http://liwc.wpengine.com/}} lexicon, which was originally designed to model psycholinguistic information from verbal communication based on specific word usages.

Hirschberg \& al. \cite{hirschberg:2005} created the Columbia-SRI-Colorado (CSC) corpus by eliciting within-speaker deceptive and non-deceptive speech. They further proposed a model based on 150 acoustic and prosodic features, and a variety of lexical features including 68 LIWC categories, filled pauses annotations, and Whissell’s Dictionary of Affect in Language (DAL) \cite{whissell1989dictionary} for extracting more emotion-related word-based features. In their experiments, using lexical and acoustic features barely outperformed the baseline, reducing the error from 39.8\% to 37.2\%. Yet, the authors managed to reduce the error further down to 33.6\% when adding speaker-dependent features such as speaker information, gender and ratio of phrases containing filled pauses and laughter.

\noindent P{\'e}rez-Rosas \& al. \cite{Perez-Rosas:2015:DDU:2818346.2820758} introduced a real-life courtroom trial dataset, including 61 deceptive and 60 truthful videos. They built a model to discriminate between liars and truth-tellers from this dataset using a variety of features such as unigrams, bigrams, manually annotated facial expressions and hand movement categories derived from the MUMIN coding scheme \cite{allwood2007mumin}, a multimodal annotation scheme for interpersonal communication focusing on the annotation of feedback, turn-management, and sequencing.

In follow-up work, Krishnamurthy \& al.~\cite{krishnamurthy2018deep} combined visual features extracted using a 3D convolutional neural (CNN) network for human action recognition \cite{3d-cnn}, which derived features from temporal and spatial dimensions by applying 3D convolutions, textual information in the form of pre-trained Word2Vec \cite{NIPS2013_5021} embeddings, audio features based on the INTERSPEECH 2013 ComParE feature set \cite{schuller2013interspeech} with noise cleaning and voice normalization, and micro-expression information from the work of P\'{e}rez-Rosas \& al.~\cite{Perez-Rosas:2015:DDU:2818346.2820758}.
Using these features as an input to a simple neural network yielded 96.1\% accuracy on the same trial dataset.

Levitan \& al. \cite{Levitan2016CombiningAL} combined and compared acoustic-prosodic, lexical, and phonotactic features, and did cross-corpus evaluation training on the Columbia Deception Corpus \cite{levitan2015individual} and testing on the Deceptive Speech Database, provided by the ComParE 2016 competition \cite{schuller2016interspeech}. They reported that their multimodal combination of representations achieved almost the same Unweighted Average Recall scores as when training and testing on the ComParE set.

The above datasets are not ideal as they are either based on recordings from staged setups where people were asked to tell the truth or to lie and their statements were recorded, or they used distant supervision, e.g.,~if a person was eventually convicted in court, that person was considered to be a liar. However, convictions could be wrong, and people probably only lie part of the time.

Many of the participants in the CLEF-2018 CheckThat! Lab Task 2, including the winning team's system \cite{Wang2018TheCT}, used a Web search engine to retrieve documents containing information relevant to the target claims, which they used as supporting evidence to decide whether the target claim was likely to be true, half-true, or false, similarly to the approach described in \cite{karadzhov2017fully}. However, they all used textual information only and none of them used acoustic features, as audio or video were not provided as part of the task.

Here, we extend prior research in deception detection to the real-world application scenario of political debates. Thanks to the efforts of fact-checking organizations, it is possible to obtain annotations for politician's statements in the context of a political discourse as true, half-true, or false. Starting with such data from the CLEF-2018 CheckThat task, which was limited to text, we performed alignment to the corresponding videos in order to recover time-stamps from the audio for the corresponding words in the speech transcripts, thus producing our multimodal dataset.

\begin{table*}[t]
\centering
\begin{tabular}{p{23mm} p{80mm} @{\hspace{5mm}} p{15mm}}
\toprule
Hillary Clinton:	& I think my husband did a pretty good job in the 1990s.	& \\
Hillary Clinton:	& I think a lot about what worked and how we can make it work again\ldots	& \\
Donald Trump:	& Well, he approved NAFTA\ldots	&  	\textbf{half-true}	\\	
\midrule
Hillary Clinton:	& He provided a good middle-class life for us, but the people he worked for, he expected the bargain to be kept on both sides. 	 &	\\
Hillary Clinton:	& And when we talk about your business, you've taken business bankruptcy six times. &	\textbf{true}	\\	
\bottomrule
\end{tabular}         
 \caption{Fragments from the 1st 2016 US presidential debate. The veracity of the check-worthy claims is shown on the right.}
 \label{fig:examples}
\end{table*}

\section{Data}
\label{sec:data}

The CT-FCC-18 corpus used in the CLEF-2018 CheckThat! Lab Task 2 \cite{clef2018checkthat:task2,nakov2018overview} contains 94 claims from three debates as a training set, and 192 claims from seven debates and speeches as a test set. The corpus includes a total of eight speakers, some of which participated in multiple debates. The information provided for each claim includes the debate or the speech it is from, the name of the speaker, the claim text as it was originally made, and whether the claim was found to be true, half-true, or false. As some of the claims heavily relied on their contexts, the lab organizers also provided manually created normalized forms of each claim. The distribution of true/half-true/false claims in the training set is 22/24/48, and it is 71/39/82 for the test set, meaning that the corpus is imbalanced. We did not use the normalized claims in our experiments since the acoustics would not match the spoken text.

Starting with the corresponding event videos, we used Kaldi\footnote{\url{http://kaldi-asr.org/}} along with the Gentle forced aligner tool\footnote{\url{http://github.com/lowerquality/gentle}} to align the speech with the text of the claim and to obtain timestamps in the audio of the starting and the ending words for each claim, thus enriching the dataset with audio. Unfortunately, this failed for 32 of the claims and we had to do the alignment manually for them. Overall, the total audio duration is 33 minutes and the average claim duration is 7 seconds.

\section{Features}
\label{sec:features}

\subsection{Textual Features}

\hspace{\parindent} \textbf{LIWC} We used as features the proportion of words in the claim that are in each of the 64 LIWC 2007 categories.

\textbf{TF.IDF} We also used TF.IDF-weighted \cite{NOdoi:10.1108/eb026526} word uni-grams, bi-grams, tri-grams, and four-grams.

\textbf{BERT} We further used the [CLS] token from the pre-trained BERT-Base, Uncased model \cite{devlin2018bert}, which yielded a 768-dimensional representation. As we had very little training data, we could not afford to perform fine-tuning as is common when using BERT.

\subsection{Acoustic Features}

\hspace{\parindent} \textbf{ComParE} We used the acoustic feature set from INTERSPEECH 2013 ComParE \cite{schuller2013interspeech}, which is a slightly modified version of the acoustic feature set used for the baseline system of the IS 2012 Speaker Trait Challenge \cite{inproceedings-is2012-speaker-trait}, which itself is the result of gradual feature set improvements and unification over previous INTERSPEECH and other audio related challenges \cite{inproceedings-is2010-paralinguistic-challenge,inproceedings-is2011-speaker-state,schuller2011avec}. In particular, we used openSMILE \cite{Eyben:2013:RDO:2502081.2502224} in order to compute the feature values for the acoustic signal for each claim. The set includes 6,373 features including psycho-acoustic spectral sharpness, voice quality, energy, spectral, and other low-level descriptors such as Mel-frequency cepstral coefficients (MFCC).
     
\textbf{i--vector} features.
The i--vector \cite{ali2015automatic} involves modeling the speech signal using a universal background model (UBM), which typically is a large Gaussian Mixture Model (GMM), trained on a large amount of data in order to represent general feature characteristics, which plays the role of a prior on how speech style looks like. The i--vector is a powerful technique that summarizes all the updates happening during the adaptation of the UBM mean components to a given utterance. All this information is modeled in a low-dimensional subspace referred to as the total variability space. In the i--vector framework, each speech utterance can be represented as a GMM super-vector. The i--vector is the low-dimensional representation of an audio recording that can be used for classification as well as for estimation purposes. In our experiments, the UBM was a GMM with 2,048 components, we used acoustic bottleneck features, the i--vectors were 600--dimensional, and there was one i--vector calculated for each claim.

\subsection{Metadata Features}

\textbf{Speaker} Finally, we included speaker information, in the form of a speaker ID, which we encoded using one-hot representation. In our experiments, this representation was always used together with and directly concatenated to the LIWC feature vectors (see above).

\begin{table}[tbh]
  \centering
 \scalebox{0.95}{
 \setlength\tabcolsep{4pt}
  \begin{tabular}{l c c c c c}
    \toprule
    \textbf{Model \& Features} & \textbf{MAE} & \textbf{MMAE} & \textbf{Acc} & \textbf{F1} & \textbf{MAR} \\
    \midrule
    BERT & \bf 0.76 & 0.78 & 43.75 & 38.63 & 39.44 \\
    i--vector & 0.77 & \textbf{0.74} & 42.71 & \textbf{40.48} &\textbf{42.92} \\
    TF.IDF n-grams & 0.81 & 0.84 & \textbf{44.79} & 33.66 & 39.37 \\
    LIWC + Speaker & 0.85 & 0.82 & 33.33 & 32.52 & 34.20 \\
    ComParE & 0.95 & 0.89 & 35.94 & 33.33 & 35.56 \\
    \midrule
    Probability avg & 0.78 & 0.78 & 40.62 & 37.14 & 38.16 \\
    Feature concatenation & 0.91 & 0.88 & 35.94 & 33.80 & 33.78 \\
    Ensemble & 0.94 & 0.88 & 31.25 & 30.97 & 32.44 \\
    \midrule
    Baseline n-gram	& 0.91	&   0.92 	&   39.57 	&   30.95	&   35.88	\\
    Baseline random	& 0.83	&   0.81 	&   35.97 	&   35.69 	&   35.89	\\    
    \bottomrule
  \end{tabular}
 }
   \caption{\label{tab:logistic_regression_experiments}Logistic regression experiments.}
\end{table}

\section {Experiments and Evaluation}
\label{sec:experiments}

\subsection{Evaluation Measures}

For evaluation, we used the evaluation measures used to rank the participants in the CLEF CheckThat! Lab task 2:
Mean Absolute Error (MAE), Macro-average Mean Absolute Error (MMAE), Accuracy, Macro-average F1, and Macro-average Recall (MAR). MAE and MMAE are calculated based on the ordering of the classes (false:0, half-true:1, true:2), taking into account that confusion between neighbouring classes is less harmful than such between more distant ones. Although MAE was the official evaluation measure at the competition, here we chose to use MAR for hyper-parameter optimization because of its higher stability for imbalanced data. Yet, we still consider MAE as the main evaluation measure.

\subsection{Logistic Regression Experiments}

We trained a logistic regression (LR) classifier using each of the above-described individual feature types. 
We addressed class imbalance by weighting the samples with weights inversely proportional to their class frequencies.
We further tuned the value of the L2 regularization parameter using leave-one-debate-out cross-validation on the training dataset. Note that this does not guarantee that there are no (near) duplicate claims by the same speaker between the debates; yet, we found such duplicates to be rare, and their effect to be further reduced due to the differences in the audio signal.

The results are shown in Table~\ref{tab:logistic_regression_experiments}. We can see that the best individual feature type is BERT, with a MAE of 0.76. It is nearly tied with i--vector, which has a MAE of 0.77, and it is also the best overall in terms of MMAE, F1, and MAR. At the third position with a MAE of 0.81 is TF.IDF n-grams, which performed best overall in terms of Accuracy. The remaining two feature types could not improve over a random baseline. Overall, both textual and acoustic features seem to be important, as we find BERT and i--vector as the top-2 based on four of our five evaluation measures.

\begin{table}[tbh]
  \centering
 \scalebox{0.95}{
 \setlength\tabcolsep{4pt}
  \begin{tabular}{l c c c c c}
    \toprule
    \textbf{Model \& Features} & \textbf{MAE} & \textbf{MMAE} & \textbf{Acc} & \textbf{F1} & \textbf{MAR} \\
    \midrule
    LIWC + Speaker & \bf 0.80 & 0.81 & 39.58 & 35.23 & 36.00 \\
    i--vector & 0.86 & \bf 0.73 & 21.35 & 15.14 & 33.12 \\
    ComParE & 0.86 & 0.83 & 41.15 & \bf 39.57 & \bf 39.83 \\
    BERT & 0.90 & 0.85 & 31.77 & 31.67 & 33.74 \\
    TF.IDF n-grams & 0.94 & 0.98 & \bf 40.62 & 25.35 & 33.24 \\
    \midrule
    \textbf{All} & \textbf{0.67} & \textbf{0.69} & \textbf{51.04} & \textbf{45.07} & \textbf{47.25} \\
    $-$ TF.IDF n-grams & 0.93 & 0.89 & 32.81 & 31.67 & 33.28 \\
    $-$ LIWC + Speaker & 0.89 & 0.83 & 33.33 & 31.19 & 35.55 \\
    $-$ i--vector & 0.86 & 0.81 & 35.94 & 35.66 & 37.12 \\
    $-$ BERT & 0.86 & 0.80 & 35.94 & 35.62 & 39.04 \\
    $-$ ComParE & 0.85 & 0.74 & 34.90 & 34.32 & 42.26 \\
    \bottomrule
  \end{tabular}
 }
 \caption{\label{tab:multi_input_nn_experiments}Neural network experiments.}
\end{table}

\noindent We further experimented with three combinations of the individual feature types. (\emph{i})~\emph{Feature concatenation:} we concatenated all features into a long feature vector, and we trained a model using this representation. (\emph{ii})~\emph{Probability avg:} we averaged the probabilities returned by the individual models, and we then made a prediction based on the highest probability. (\emph{iii})~\emph{Ensemble:} We trained an ensemble by adding a logistic regression meta-classifier on top of the predictions of the individual models. In order to tune the meta-classifier's L2 parameter, we recorded the posterior probability distribution for each base classifier for each training example using leave-one-out debate cross-validation, and then we performed another cross-validation for the meta-classifier on these recorded scores. At the end, we retrained the individual base classifiers and the meta-classifier once again, but using the recorded probabilities for the entire dataset.

The evaluation results are shown in the middle of Table~\ref{tab:logistic_regression_experiments}. We can see that none of the combinations could improve over the top-2 systems: this is probably due to the small number of training examples, which made it hard for the combined models to handle the increased number of parameters.

\begin{figure*}[t]
  \centering
  \includegraphics[width=\linewidth]{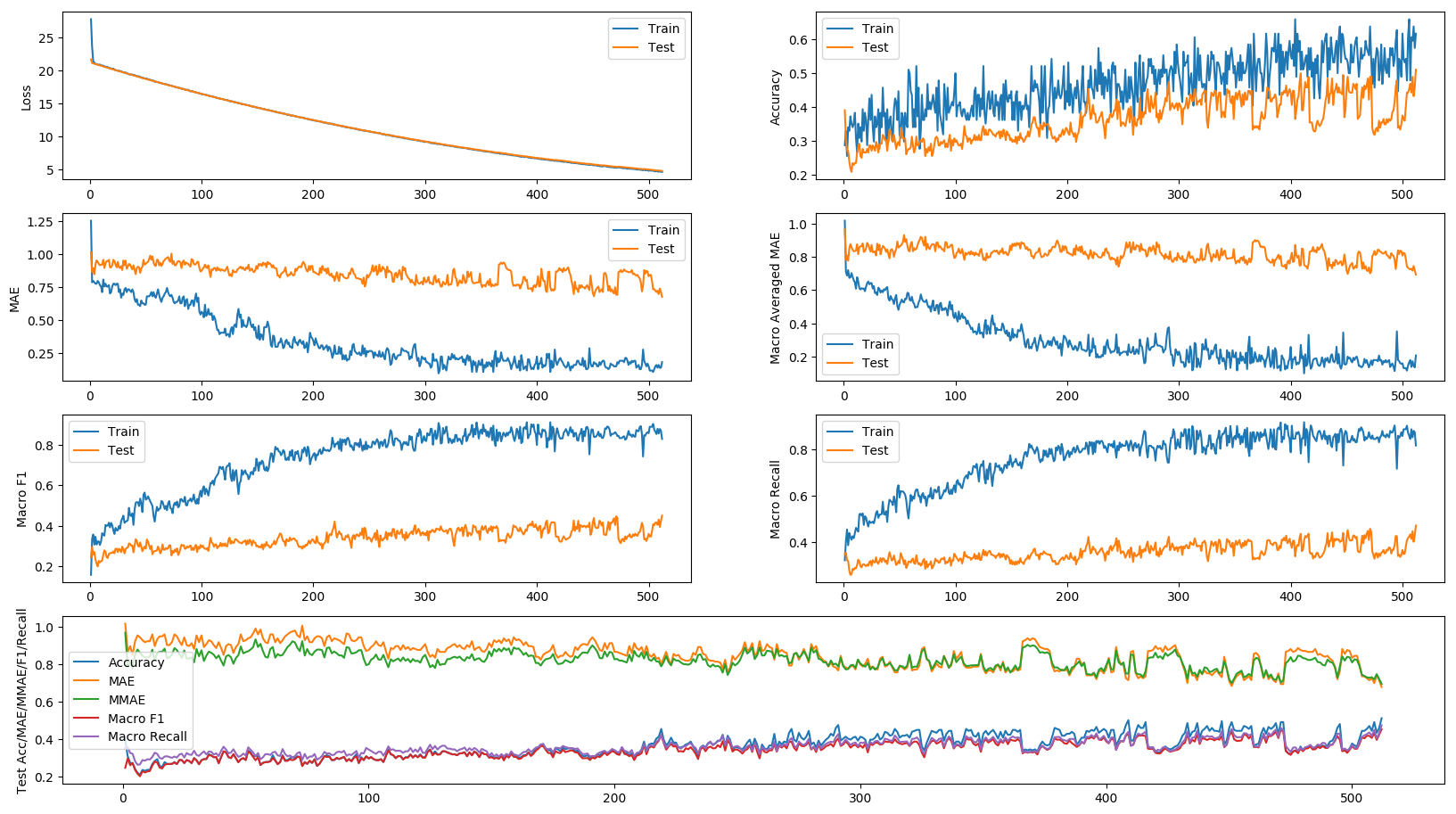}
  \caption{Training history for the full multi-input neural network.}
  \label{fig:performance_graphics}
\end{figure*}

\subsection{Neural Network Experiments}

\begin{figure}[ht]
  \centering
  \includegraphics[width=\linewidth]{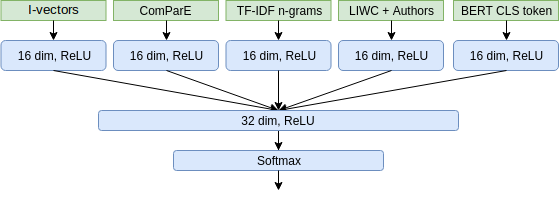}
  \caption{The architecture of our multimodal neural network.}
  \label{fig:arch_diagram}
\end{figure}

Next, we experimented with a multi-input feed-forward neural network, shown in Figure~\ref{fig:arch_diagram}. It takes all feature types at once and then for each one there is a separate fully connected hidden layer of size 16, the outputs of which are concatenated and fed into another fully connected hidden layer of size 32, followed by an output layer. We used rectified linear units (ReLU) for activation for all connections, except for the final prediction, for which we used softmax. We trained the network for 512 epochs using stochastic gradient descent without momentum, a learning rate of 0.005, and a dropout retention rate of 0.5. As in the logistic regression experiments above, we put weights to the cross-entropy loss so that individual examples are weighed inversely proportionally to the class imbalance.

\begin{table}[ht]
 \scalebox{0.90}{
 \setlength\tabcolsep{2.5pt}
  \begin{tabular}{l@{ }c c@{ }c c c}
    \toprule
    \textbf{System} & \textbf{MAE} & \textbf{MMAE} & \textbf{Acc} & \textbf{F1} & \textbf{MAR} \\
    \midrule
    Multi-input neural network & \textbf{0.67} & 0.69 & \textbf{51.04} & \textbf{45.07} & \textbf{47.25} \\
    \cite{Wang2018TheCT} Copenhagen & 0.71 & \textbf{0.67} &  43.17	&   40.08	&  45.02	\\
    FACTR & 0.91	&   0.93	&   41.01	&   32.36	&   36.84	\\
    \cite{Ghanem:2018} UPV--INAOE--Autoritas & 0.95	&   0.97	&   38.85	&   26.13	&   34.03	\\ 
    \midrule
    Baseline n-gram	& 0.91	&   0.92 	&   39.57 	&   30.95	&   35.88	\\
    Baseline random	& 0.83	&   0.81 	&   35.97 	&   35.69 	&   35.89	\\ 
    \bottomrule
  \end{tabular}
  }
    \caption{\label{tab:checkthat_2018_comparison}Comparison to the top-3 systems at the CLEF-2018 CheckThat! Lab Task 2.}
\end{table}

Figure~\ref{fig:performance_graphics} shows the loss and the performance for our full multi-input neural model on training for the different evaluation measures after each epoch. Note that there is a higher than usual variance in the values, as well as fairly small improvement for some of the measures on the test data, which we attribute to the rather small number of training examples. 

Table~\ref{tab:multi_input_nn_experiments} reports the evaluation results for our multi-input neural network model using the same architecture and hyper-parameters as in Figure~\ref{fig:performance_graphics}. The top half of the table shows the results when only one feature type is given as an input and the rest are set to zero. As in the logistic regression experiments, we see i--vector at the second-best position. However, this time the best feature type is LIWC + Speaker rather than BERT, which can be due to some information being lost when passing the BERT representation via two hidden layers as opposed to using it directly.

The top line of the bottom half of Table~\ref{tab:multi_input_nn_experiments} shows the results when using the full model with all features. We can see that this yields sizable improvements over using any individual feature type, which is in contrast to our combination experiments for the logistic regression experiments above. The following lines show ablation experiments where we leave one feature type out. Interestingly, even though the TF.IDF-weighted word n-gram features perform worse when used in isolation, they cause the largest drop for almost all evaluation measures, which suggests that they add some information that is complementary to what the remaining features already provide. The following two feature types whose removal causes the highest drop in performance are LIWC + Speakers and i--vector, which should not be surprising as these are the two best individual feature types from the top of the table.

\begin{table}[tbh]
    \centering
    \begin{tabular}{cc|c|c|c|l}
        \cline{3-5}
        & & \multicolumn{3}{ c| }{\textbf{Predicted Label}} \\ \cline{3-5}
        & & False & Half-True & True \\ \cline{1-5}
        \multicolumn{1}{ |c  }{\multirow{3}{*}{\textbf{True Label}} } &
        \multicolumn{1}{ |c| }{False} & 66 & 12 & 4    \\ \cline{2-5}
        \multicolumn{1}{ |c  }{}                        &
        \multicolumn{1}{ |c| }{Half-True} & 16 & 14 & 9    \\ \cline{2-5}
        \multicolumn{1}{ |c  }{}                        &
        \multicolumn{1}{ |c| }{True} & 32 & 21 & 18     \\ \cline{1-5}
    \end{tabular}
    \caption{\label{tab:conf_matrix_nn}Confusion matrix for the neural network.}
\end{table}

Next, Table~\ref{tab:checkthat_2018_comparison} compares our results to those for the top-3 systems from CLEF-2018 CheckThat! Lab Task 2. We can see that our model achieves sizable improvements over the best system for all evaluation measures except for MMAE, thus establishing a new state of the art (thanks to the acoustic information). Note that some of these top-3 systems perform close to or fall behind the baselines, which demonstrates the difficulty of the task.

Finally, Table~\ref{tab:conf_matrix_nn} shows a confusion matrix for our full neural model. We can see that the model tends to confuse the True class with the False and the Half-True classes, especially with the latter. Yet, this bias in the confusion matrix can be easily explained with the class imbalance in the data.

\section{Conclusion and Future Work}
\label{sec:conclusion}

We have presented the first multimodal dataset for fact-checking the claims made in a political debate, which we release to the research community, hoping to help advance research in multimodal deception detection. We further developed a multimodal deep-learning architecture, which combines textual and acoustic information, yielding sizable improvements over the state of the art for the CLEF-2018 CheckThat! Lab Task 2. Notably, we use no external information, which would not be available when a brand new claim is made for the first time. Thus, we believe that our framework is well tailored for the real world.

In future work, we plan to extend our framework to use information from the video \cite{Perez-Rosas:2015:DDU:2818346.2820758}. We further want to extend the dataset with more debates and speeches. Finally, we would like to experiment with languages other than English.

\section{Acknowledgments}

This research is part of the Tanbih project,\footnote{\url{http://tanbih.qcri.org/}} which aims to limit the effect of ``fake news'', propaganda and media bias by making users aware of what they are reading. The project is developed in collaboration between the Qatar Computing Research Institute (QCRI), HBKU and the MIT Computer Science and Artificial Intelligence Laboratory (CSAIL).

This research is also partially supported by Project UNITe BG05M2OP001-1.001-0004 funded by the OP ``Science and Education for Smart Growth'' and co-funded by the EU through the ESI Funds.

\bibliographystyle{IEEEbib}
\bibliography{refs}

\end{document}